\documentclass[conference]{IEEEtran}
\IEEEoverridecommandlockouts
\usepackage{cite}
\usepackage[utf8]{inputenc}
\usepackage{amsmath,amssymb,amsfonts}
\usepackage{graphicx}
\usepackage{textcomp}
\usepackage{caption}
\usepackage{siunitx}
\usepackage{xcolor}
\usepackage{mathptmx,mathtools,mathrsfs}
\usepackage{float}
\usepackage[table]{xcolor}
\definecolor{lightgreen}{RGB}{200,255,200}
\usepackage[ruled,lined,linesnumbered,noend]{algorithm2e}
\usepackage{tabularx}
\usepackage{makecell}
\usepackage{todonotes}

\def\BibTeX{{\rm B\kern-.05em{\sc i\kern-.025em b}\kern-.08em
    T\kern-.1667em\lower.7ex\hbox{E}\kern-.125emX}}
\begin{document}

\title{Low-Energy Reduced \mbox{RISC-V} Instruction Subset Processor
for Tsetlin Machine Inference at the Edge}

\author{\IEEEauthorblockN{Chanda Gupta$^1$, Sanidhya Bhatia$^1$, Shaurya Priyadarshi$^1$, Himani Panwar$^1$, Rishad Shafik$^2$ and Sudip Roy$^1$}

\IEEEauthorblockA{$^1$\textit{CoDA Laboratory}, \textit{Indian Institute of Technology Roorkee}, Roorkee, India\\
$^2$\textit{Microsystems Research Group}, \textit{Newcastle University}, Newcastle upon Tyne, UK\\
\{chanda\_g, s\_bhatia, s\_priyadarshi, h\_panwar, sudip.roy\}@cs.iitr.ac.in, rishad.shafik@newcastle.ac.uk}
}
\maketitle
\begin{abstract}
Tsetlin Machine (TM) is a logic-based machine learning approach that relies on simple bitwise operations and finite-state automata, which makes it attractive for edge AI deployments. Recent work has focused on co-processor and accelerator designs based on Tsetlin Machines (TMs). Although these designs achieve high performance, they typically depend on tightly coupled interfaces, microcode-style programming, and external host processors, limiting flexibility and ease of programming.
In this work, we present a domain-specific \mbox{RISC-V} microprocessor architecture and design flow tailored for TM inference. Leveraging the modular structure of RISC-V, we design a reduced instruction subset processor that retains programmability while targeting improved performance and lower energy consumption for TM workloads. Instruction profiling is employed to guide instruction reduction, followed by datapath and control path simplifications tailored to TM inference.
Both the baseline RV32IM core and the proposed reduced core are evaluated across multiple datasets and compared with Binarized Neural Networks (BNNs), which serve as a hardware-efficient baseline due to their reliance on bitwise operations during inference. 
Results show that TM achieves comparable or higher accuracy (e.g., up to 88.18\% on CIFAR-2 compared to 60.0\% for BNN) while reducing execution time by up to 98\% across multiple datasets. Furthermore, the proposed design achieves an average $29.7\times$ reduction in energy consumption, demonstrating its effectiveness for programmable and efficient edge AI systems.
\end{abstract}


\section{Introduction}
The rapid growth of edge computing in domains such as the Internet of Things (IoT), smart sensing, and autonomous systems has created a strong demand for energy-efficient and low-complexity processors capable of performing edge \mbox{Artificial} Intelligence (AI) inference. To meet these requirements, researchers have
explored specialized hardware architectures for accelerating machine learning workloads~\cite{2024_ISCAS_Hsiao,2025_CSI_Tunheim}. 
One common approach is the use of Application-Specific Integrated Circuits (ASICs), which implement fixed-function hardware tailored to specific algorithms~\cite{2020_ISSCC_dean}.
Processors like Apple's Neural Engine and Google's Tensor Processing Unit (TPU) have shown high-performance AI inference using ASIC-based designs. However, these designs usually lack the flexibility needed to accommodate changing machine learning models because they are optimized for a limited set of operations.
To overcome this limitation, Application-Specific Instruction-set Processors (ASIPs) extend conventional processors with customized instruction sets and specialized datapaths while preserving programmability. Dedicated hardware support for signal processing and low-power computation is offered by embedded systems like the Texas Instruments MSP430 microcontroller and its Low Energy Accelerator (LEA)~\cite{2025_CSI_Tunheim}.
\mbox{Despite} these benefits, ASIP-based systems can increase
design complexity and require
careful hardware-software co-design.

The idea of open hardware has recently gained traction, allowing researchers to create more accessible and customizable CPU architectures~\cite{2025_ISTM_Zeng,riscv}. 
Among these, the open-source RISC-V instruction set architecture (ISA) has gained
a lot of interest because of its modular design and support
for custom extensions, which allow for the creation of domain-specific processors 
while maintaining 
programmability.
Neural network models, such as Transformer architectures and Convolutional Neural Networks (CNNs), are the main focus of the majority of current domain-specific processor research [3].
However, these models rely on compute-intensive
 operations and large memory footprints, resulting in
complex processor implementations and high computational
costs.

Tsetlin Machine (TM) has recently emerged as a novel machine learning paradigm based on propositional logic rather than arithmetic computation~\cite{granmo2018tsetlin}. Built on Boolean expressions and simple decision rules, 
TM models have been shown in prior literature to achieve low computational complexity, high interpretability, and significant energy efficiency~\cite{2025_IWASI_Duan,2026_PR_Tarasyuk}.
These characteristics make TM well-suited for resource-constrained edge AI systems. However, efficiently mapping such algorithms onto programmable processor architectures remains an open challenge. Our hypothesis is that such an approach can enable processor architectures with ultra-low energy consumption, fast operation, and reduced overall system overhead.
For comparative analysis, we consider a neural network-based characterization using Binarized Neural Networks (BNNs)~\cite{2016_BNN_courbariaux}, which utilize binary weights and activations for inference, and perform comprehensive evaluations.

We make the following key contributions:
\begin{itemize}
\item Design of a programmable and high-performance domain-specific \mbox{RISC-V} architecture tailored for Tsetlin Machine inference.
   
\item Extensive validation of the proposed architecture across multiple TM workloads, demonstrating the trade-offs between performance, accuracy, and programmability.
   
\item Comparative analysis with Binarized Neural Networks (BNNs) on \mbox{RISC-V} processors to evaluate performance and energy efficiency.
  
\end{itemize}

The remainder of the paper is organized as follows.
Sec.~\ref{section:BACKGROUND_RELATED_WORKS} presents the background of the proposed work.
Section~\ref{section:proposed} describes the proposed method,
 simulation results are provided in Sec.~\ref{section:SIMULATION_RESULTS} and finally, Sec.~\ref{sec:CONCLUSION} concludes the paper.


\section{Background}
This section presents background on the TM model, BNNs and \mbox{RISC-V} for domain-specific architectures.
\label{section:BACKGROUND_RELATED_WORKS}

\begin{figure}[!t]
    \centering
    \includegraphics[width=0.88\linewidth]{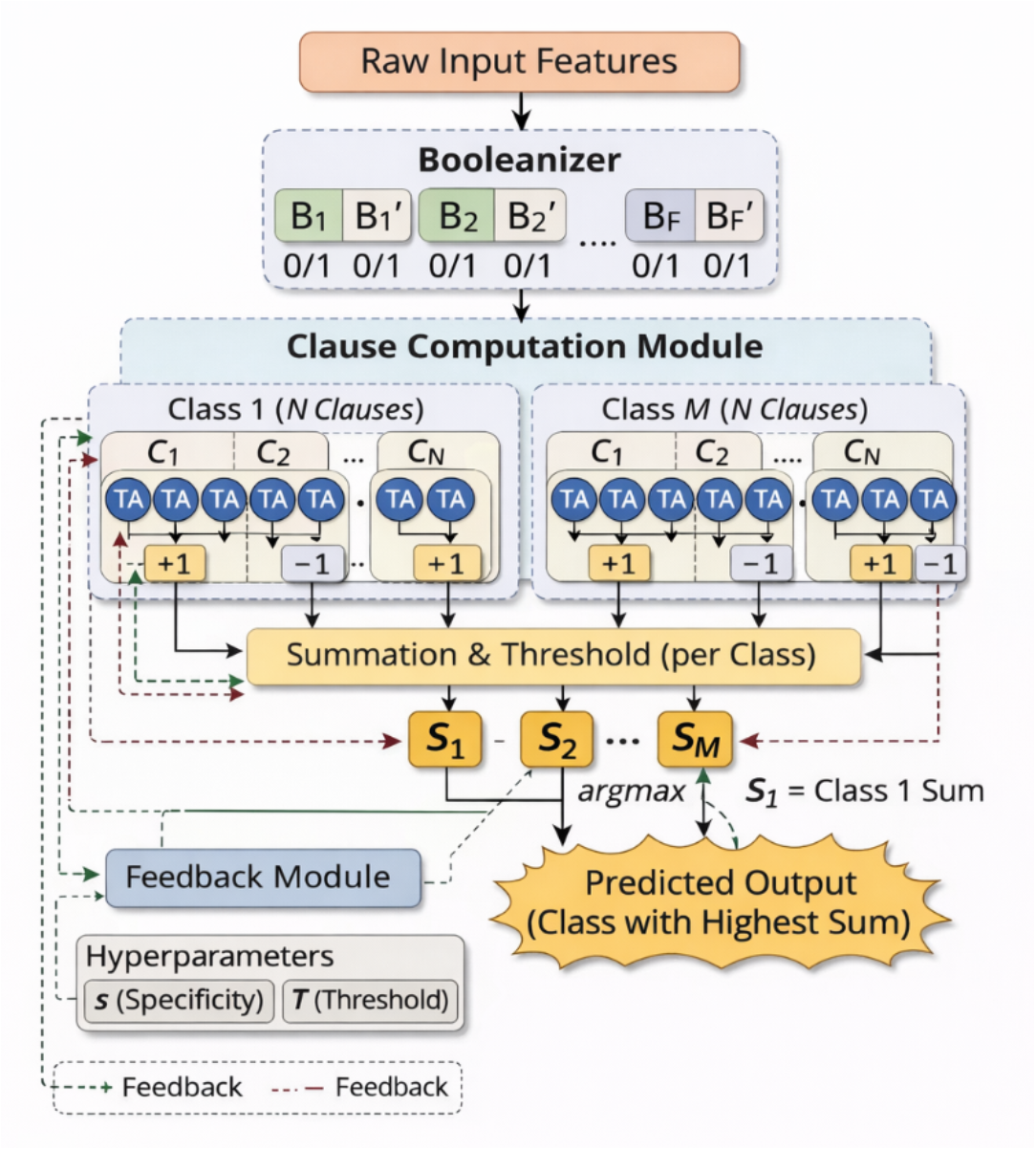}
    \caption{Block diagram of  a multiclass Tsetlin machine.}
    \label{fig:TM}
\end{figure}

\subsection{Tsetlin Machine Model and BNNs}

Lightweight machine learning models are increasingly explored for edge intelligence due to their reduced computational requirements.
Among these, Binarized Neural Networks (BNNs) and Tsetlin Machines (TM) have drawn interest as effective substitutes for traditional deep learning models.
Tsetlin Machine (TM) is an logic-based machine learning model that uses propositional logic for pattern recognition tasks~\cite{granmo2018tsetlin}.
Its core components include Tsetlin Automata (TA) organized into conjunctive clauses, along with summation, threshold, and feedback modules.
BNNs are derived from conventional deep neural networks by constraining weights and activations to binary values.
They employ multi-layer architectures and backpropagation-based training, replacing costly arithmetic operations with efficient bitwise operations.
This significantly reduces computational complexity and memory usage, making BNNs suitable for resource-constrained edge environments.
For multi-class classification, we employ the Multiclass TM, where each class is represented by a set of clauses. Since most real-world tasks involve multiple classes, the terms TM and Multiclass TM are used interchangeably in this work. As illustrated in Fig.~1, a Multiclass TM consists of $M$ classes, each containing $N$ clauses. Each clause is constructed from a group of TAs, where the number of automata equals twice the number of Boolean input features to account for both inputs and their negations. A Booleanizer converts raw input features into binary values and their complements, which serve as inputs to the clause computation module. For each datapoint, the clause computation stage produces a 1-bit output per clause based on the collective decisions of the associated TAs. Each class contains an even number of clauses with alternating polarities of $+1$ and $-1$, indicating whether a clause supports or opposes the classification outcome. The clause outputs are aggregated through a summation process to compute a class sum for each class, and an $\arg\max$ operation selects the class with the highest sum as the predicted output. 
Using straightforward logical procedures, this voting-based system allows for effective and comprehensible classification. The threshold $T$ and the specificity parameter $s$ are two important hyperparameters in TM. While $s$ establishes the likelihood of using literals in clauses, the $T$
regulates the number of clauses that take part in voting.

\subsection{RISC-V for Domain-Specific Architectures}
An open-source, modular, extensible platform for processor design is offered by \mbox{RISC-V}, which enables designers to set up cores for domain-specific applications~\cite{riscv}. 
A basic standardized base instruction set, such as RV32I, is defined by the \mbox{RISC-V} ISA and can be expanded or decreased. 
It provides fine-grained control over memory management, execution units, and the processor pipeline to optimize deterministic machine learning workloads. 
In addition, rapid development and deployment are supported by established toolchains of \mbox{RISC-V}, including open-source cores and GCC-based compilers.
Furthermore, efficient mapping of application-specific workloads onto hardware is made possible by the ability to customize both the instruction subset and microarchitecture, which minimizes needless computational cost.Therefore, \mbox{RISC-V} serves as an efficient hardware platform for logic-based, lightweight machine learning models such as the Tsetlin Machine.
Scalable edge AI deployments require a balance between programmability and hardware efficiency, which is made possible by such domain-specific customization.

\begin{figure}[!t]
    \centering
    \includegraphics[width=0.97\linewidth]{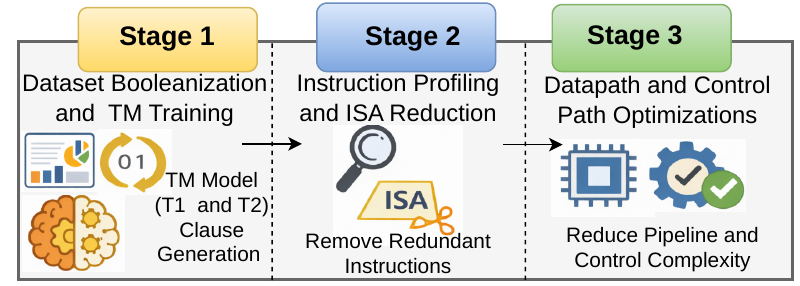}
    \caption{Proposed design flow for reduced RISC-V architecture.}
    \label{fig:workflow}
\end{figure}

\section{Proposed Reduced RISC-V Design Flow}
\label{section:proposed}
This section presents the proposed  RISC-V design flow for Tsetlin Machine inference, organized into three stages.
\subsection{Overview of the Proposed Design Flow }
The proposed design flow consists of three stages, as shown in Fig.~\ref{fig:workflow}. Stage 1 performs dataset Booleanization and Tsetlin Machine (TM) model generation. Stage 2 focuses on instruction profiling to enable ISA reduction along with initial datapath and control path simplifications. Stage 3 applies further refinement to the datapath and control logic to reduce complexity and improve efficiency for TM inference on the \mbox{RISC-V} processor.

{\color{black}
\subsection{Tsetlin Machine Model Generation} }
We implemented multiple inference strategies on \mbox{RISC-V} cores for comprehensive evaluation. Two Tsetlin Machine approaches (T1 and T2) were used, differing in clause storage and evaluation, along with a baseline BNN inference for fair comparison due to its similar bitwise operations.

1) \textbf{Vanilla TM inference (T1):}
Vanilla TM inference refers to the standard clause evaluation process without any optimization techniques.
In this strategy, each clause is evaluated by sequentially checking every literal for inclusion and then comparing it against the input literal vector. After clause evaluation, the vote sum is computed sequentially for each class, and the class with the highest vote sum is selected as the output. The literal states (include/exclude) are stored in memory for each clause as boolean values. This procedure is formalized in 
\textbf{Algorithm~\ref{ALGO:Stand_TM}}.
At lines $1$–$8$, the algorithm iterates over each class and clause, evaluating useful clauses by checking literal inclusion and updating the clause output through logical operations, while non-useful clauses are skipped to avoid unnecessary computation. Subsequently, lines $9$–$16$ compute the vote total for each class by incrementing or decrementing votes based on clause polarity. Finally, at line $17$, the class with the maximum vote value is selected as the predicted output. It is implemented in C++, and the \mbox{RISC-V} GNU toolchain~\cite{gnu} is used for cross-compilation. 

2) \textbf{Modified TM inference (T2):}
To improve efficiency, we implemented a second inference strategy inspired by~\cite{redress}, in which each clause stores only the indices of literals that are actually included, encoded as half-words (16-bit values).
Only these indices are accessed and assessed during inference. The number of memory accesses, memory usage, and logical comparisons needed for clause evaluation are all greatly decreased by this sparse clause structure. For example, even if only 10 literals are active in a sentence containing 128 literals, the usual technique needs keeping 128 inclusion flags, each taking up a byte (in C++ Inference). The improved version, on the other hand, only needs to hold ten indices, each of which usually requires two bytes, greatly lowering the memory footprint in these situations. The improved strategy ensures functional equivalency with the traditional approach while preserving the semantics of clause evaluation in spite of this compression.
This procedure is formalized in 
\textbf{Algorithm~\ref{ALGO:Opti_TM}}. 
At lines $1$–$9$, the algorithm iterates over each class and clause, evaluating only the literals whose indices are stored in memory. If a clause contains included literals, the clause output is initialized and updated by sequentially checking the referenced literal values; otherwise, the clause output is set to zero to avoid unnecessary computation. Subsequently, lines $10$–$17$ compute the vote total for each class by adjusting votes based on clause polarity. Finally, at line $18$, the class with the maximum vote value is selected as the predicted output. 

\begin{algorithm}[t!]
{
\small
\caption{\textbf{Vanilla TM inference (T1)}}
\SetAlgoLined
\label{ALGO:Stand_TM}
\KwIn{literal\_values[], temp\_is\_included[][][], useful\_clause[][],
      max\_classes, max\_clauses, max\_literals, temp\_variable}
\KwOut{predicted class $result$}

\ForEach{class $i$}{
    \ForEach{clause $j$}{
        \eIf{useful\_clause[$i$][$j$]}{
            $clause\_output[i][j] \gets 1$\;
            \For{$k \gets 0$ \KwTo $max\_literals - 1$}{
                $clause\_output[i][j] \gets clause\_output[i][j] \wedge 
                ((\neg temp\_is\_included[i][j][k]) \vee literal\_values[k])$\;
            }
        }{
            $clause\_output[i][j] \gets 0$\;
        }
    }
}

\ForEach{class $i$}{
    $votes[i] \gets max\_clauses$\;
    \ForEach{clause $j$}{
        \If{$clause\_output[i][j] = 1$}{
            \eIf{$j < temp\_variable$}{
                $votes[i] \gets votes[i] + 1$\;
            }{
                $votes[i] \gets votes[i] - 1$\;
            }
        }
    }
}

$result \gets \arg\max_i \; votes[i]$\;
}
\end{algorithm}


\begin{algorithm}[t!]
\small
{
\caption{\textbf{Modified TM inference (T2)}}
\label{ALGO:Opti_TM}
\KwIn{literal\_values[], included\_literal\_offsets[][][], 
      included\_literal\_count[][], max\_classes, max\_clauses, temp\_variable}
\KwOut{predicted class $result$}

\ForEach{class $i$}{
    \ForEach{clause $j$}{
        \eIf{$included\_literal\_count[i][j] > 0$}{
            $clause\_output[i][j] \gets 1$\;
            \For{$k \gets 0$ \KwTo $included\_literal\_count[i][j] - 1$}{
                $literal\_index \gets included\_literal\_offsets[i][j][k]$\;
                $clause\_output[i][j] \gets clause\_output[i][j] \wedge 
                literal\_values[literal\_index]$\;
            }
        }{
            $clause\_output[i][j] \gets 0$\;
        }
    }
}

\ForEach{class $i$}{
    $votes[i] \gets max\_clauses$\;
    \ForEach{clause $j$}{
        \If{$clause\_output[i][j] = 1$}{
            \eIf{$j < temp\_variable$}{
                $votes[i] \gets votes[i] + 1$\;
            }{
                $votes[i] \gets votes[i] - 1$\;
            }
        }
    }
}

$result \gets \arg\max_i \; votes[i]$\;
}
\normalsize
\end{algorithm}


3)~\textbf{Binarized Neural Networks (BNNs):}
To ensure a consistent and fair comparison between TM and BNN inference, a baseline BNN inference approach was implemented in C++ and cross-compiled in a similar manner. 
We use BNNs as a hardware-efficient baseline due to their bitwise operation-based inference.
In this approach, inference is performed sequentially across layers, where each neuron evaluates the binary outputs from the previous layer against the corresponding binary weights. Matching pairs contribute positively to an accumulation sum, while mismatches reduce it, effectively realizing XNOR-based comparisons followed by integer accumulation. The accumulated value is then evaluated against a zero threshold to determine the neuron activation. This process continues until the final layer produces class scores, and the class with the highest score is selected as the predicted output. 
\subsection{Software Compilation Flow and Model Deployment}
Tsetlin Machine models were trained using the TMU (Tsetlin Machine Unified) Python library~\cite{tmu}. Multiple datasets
and clause configurations were used to evaluate performance
across diverse conditions.
BNN models were trained using Larq, a TensorFlow framework for BNNs, and model weights were extracted. The extracted clauses and weights were strategically placed in the data-memory (DMEM)  alongside the test input before simulation. Inference logic was
implemented in C++, cross-compiled using the same \mbox{RISC-V} toolchain, and converted into instruction memory using
the same assembly flow. This ensured a consistent and fair comparison between TM and BNN inference in terms of
memory structure, control flow, and execution overhead.

\subsection{Instruction Profiling}
\label{subsec:instruction_profiling}
Instruction profiling was performed by analyzing the assembly generated from TM inference to identify frequently used instructions. In the absence of function calls in the \mbox{RISC-V} simulation, stack behavior is emulated by initializing the stack pointer and assigning memory addresses to parameters and local variables. Pseudo-instructions are converted into base \mbox{RISC-V} ISA instructions for compatibility with the reduced instruction subset. The conversions are as follows:

\begin{enumerate}
\renewcommand{\labelenumi}{(\roman{enumi})}
\setlength{\itemsep}{0pt}
\setlength{\parskip}{0pt}
\setlength{\labelwidth}{1.2cm}
\setlength{\leftmargin}{1.4cm}

\item \texttt{li x, imm $\rightarrow$ addi x, x0, imm}
\item \texttt{mv x, y $\rightarrow$ addi x, y, 0}
\item \texttt{beqz x, label $\rightarrow$ beq x, x0, label}
\item \texttt{snez x, y $\rightarrow$ sltu x, x0, y}
\item \texttt{zext.b x, y $\rightarrow$ andi x, y, 0xff}
\item \texttt{bge x, y, label $\rightarrow$ blt y, x, label}
\item \texttt{nop $\rightarrow$ add x0, x0, x0}

\end{enumerate}

\begin{figure}[!t]
    \centering
    \includegraphics[width=0.9999\linewidth]{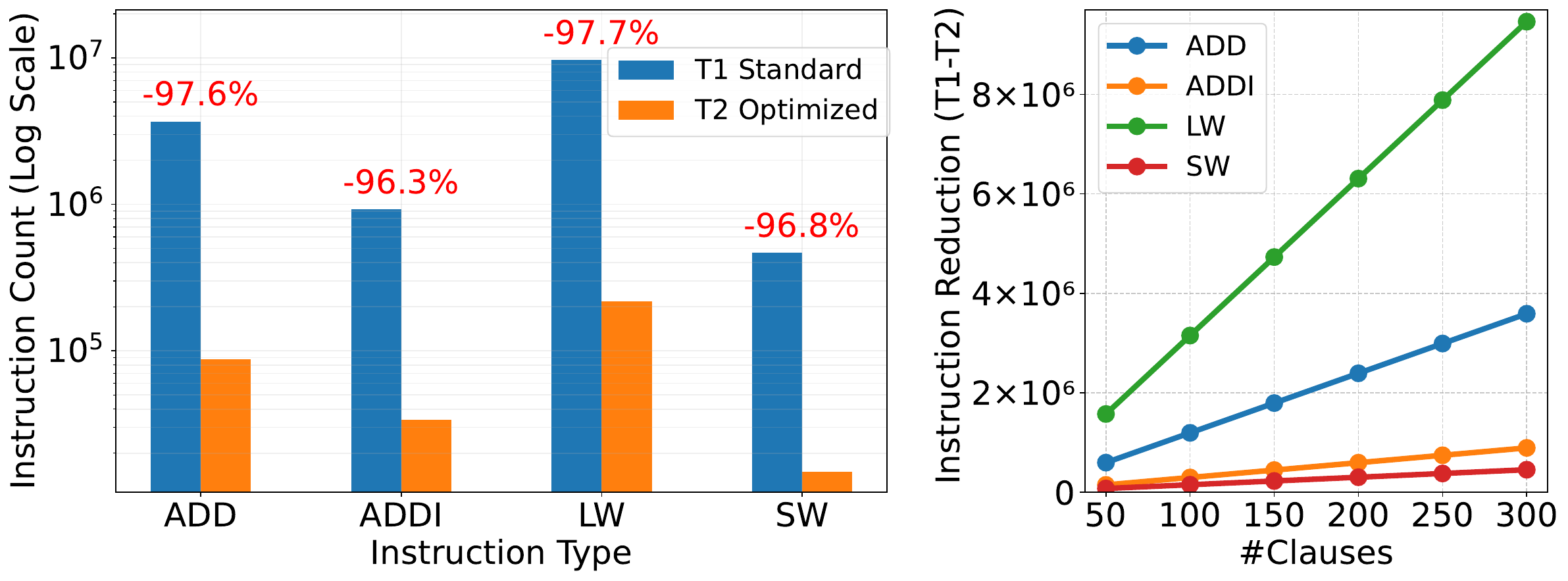}
    \caption{(a) Instruction count comparison between T1 and T2, and (b) absolute instruction reduction trend.}
    \label{fig:pareto}
\end{figure}

The impact of reducing high frequency instructions is shown in Fig.~\ref{fig:pareto}(a), demonstrating significant instruction count reduction using the modified approach. A subset of instructions contributes most to this reduction. The trend of absolute reduction with increasing clauses is illustrated in Fig.~\ref{fig:pareto}(b).

{\color{black}
\subsection{Domain-Specific Design and Optimization}}
We utilised two versions of a single-core \mbox{RISC-V} processor from the UltraEmbedded open-source GitHub repository~\cite{risc-v_repo}.
Both cores were synthesized and simulated using Xilinx Vivado, enabling us to accurately capture instruction execution behavior and analyze resource utilization.

\subsubsection {Original RISC-V Core (R1)}
\begin{figure}[!b]
    \centering
    \includegraphics[width=0.85\linewidth]{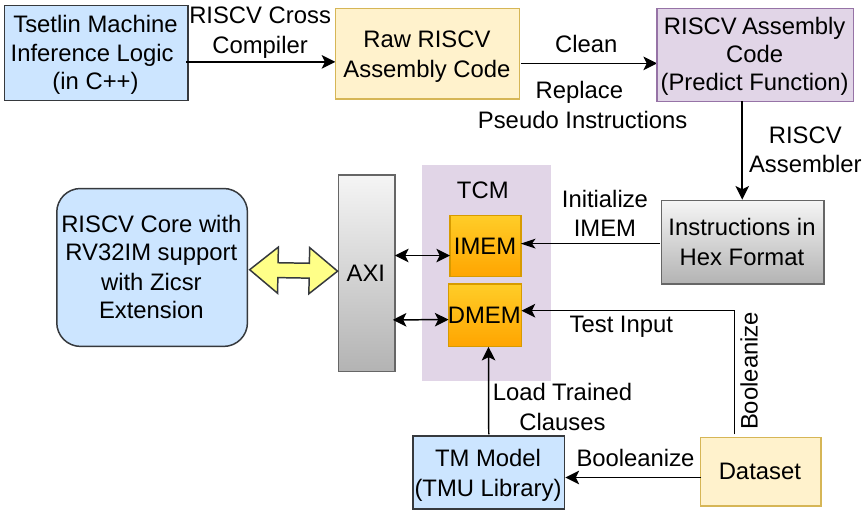}
    \caption{Hardware-software co-design approach for domain-specific RISC-V design.}
    \label{fig:hardsoftco}
\end{figure}

The original \mbox{RISC-V} core is a full RV32IM implementation, supporting the RV32I base instruction set along with the multiplication division extension (M) and the Control and Status Register (ZICSR) extensions totaling  59 instructions. It follows a classic 5-stage pipeline (Fetch, Decode, Execute, Memory, Writeback) with support for result forwarding, configurable pipeline depth, and privilege modes (user, supervisor, and machine).
The ZICSR extension
enables access to control and status registers, while memory
can be interfaced via instruction/data caches or tightly coupled
memories (TCMs) using an Advanced eXtensible Interface (AXI) bus interface. 
While Fig.~\ref{fig:workflow} presents the overall three-stage design methodology, Fig.~\ref{fig:hardsoftco} illustrates the implementation workflow and system architecture used for executing TM inference on the RISC-V processor.
In our setup, the core was connected to a tightly coupled
instruction and data memories via the AXI interface. 
The compiled inference code was loaded into instruction memory,
while clause data, input features, and output buffers
placed in data memory as shown in Fig. \ref{fig:hardsoftco}. 

\begin{figure*}[htbp]
\centering
\includegraphics[width=0.76\textwidth]{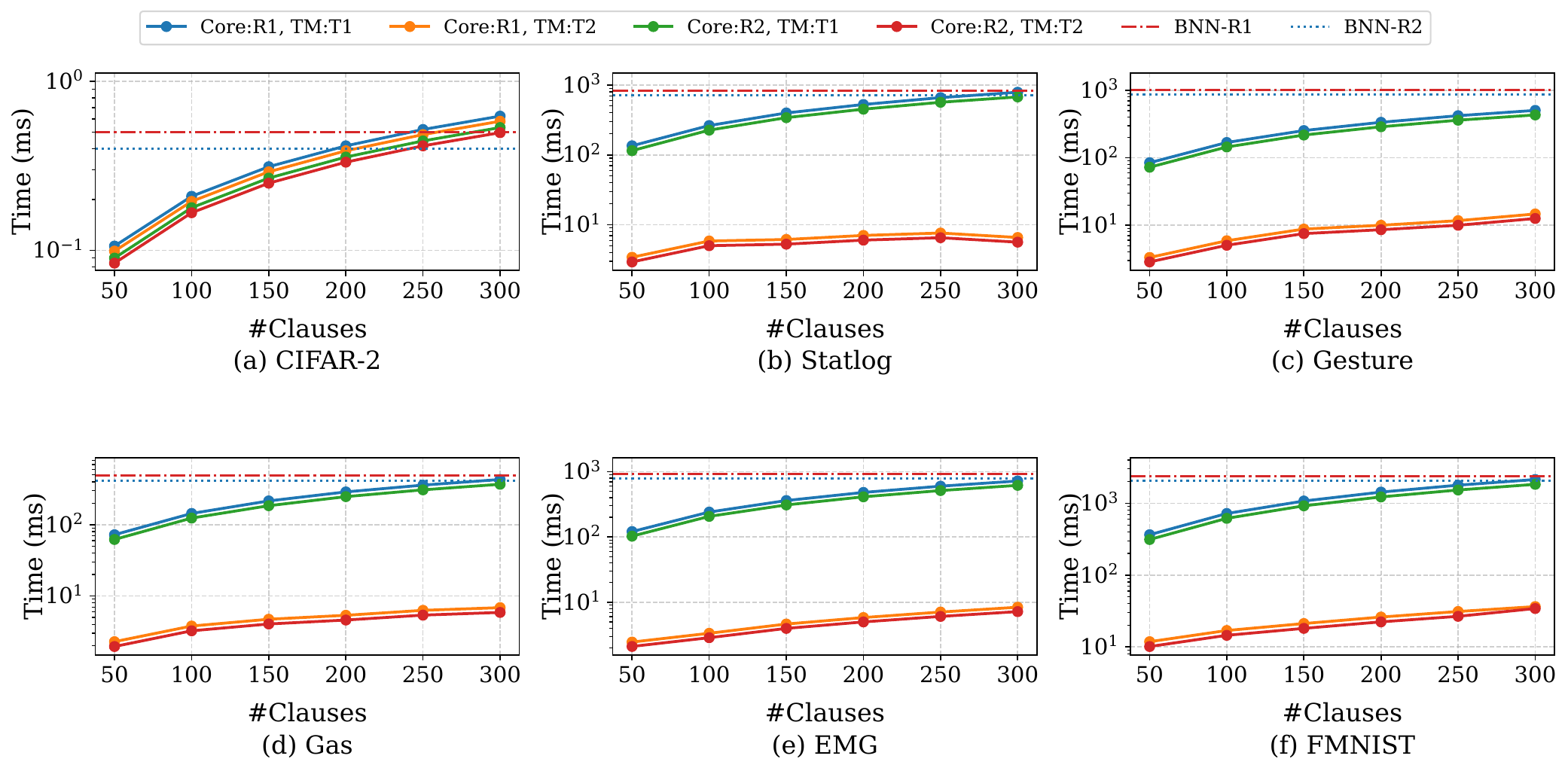}
    \caption{Execution time (in ms) across multiple inference scenarios for datasets.}
\label{fig:execution}
\end{figure*}

{\color{black}
\subsubsection {Design Modified \mbox{RISC-V} Core (R2)}

The modified core is a simplified variant that supports only the instructions required for TM and BNN inference, reducing control complexity and unnecessary hardware. Profiling (Section~\ref{subsec:instruction_profiling}) identifies frequently used instructions and critical datapath and control paths. Optimization is then focused on these key components, followed by revisiting the design to evaluate improvements in efficiency and performance.} 
To strip down the core, we used the following strategy:
\begin{enumerate}
    \item[a.] remove instruction definitions and flags,
    \item[b.] prune decode and execution logic for those instructions,
    \item[c.] simplify control and data paths.
\end{enumerate}
The core supported a total of 27 instructions after reduction. 

\begin{itemize}
    \item Base ISA Instructions (21): 
\noindent\texttt{andi,\! addi,\! xori,\! slli,\! add,\! lui,\! sltu,\! or,\! and,\! jal,\! beq,\! bne,\! blt,\! lw,\! lbu,\! lhu,\! sb,\! sw,\! ecall,\! ebreak,\! eret}
    \item ZICSR Extension Instructions (6): \noindent\texttt{csrrw,\! csrrs,\! csrrc,\! csrrwt,\! csrrsi,\! csrrci}
\end{itemize}

\section{Experimental Results}
\label{section:SIMULATION_RESULTS}
The designs were synthesized in Xilinx Vivado targeting the Zynq-7000 ZC702 Evaluation Board, and the on-chip power consumption of both cores was analyzed. The original \mbox{RISC-V} core (R1) exhibited a total on-chip power consumption of \SI{0.138}{\watt}, while the modified \mbox{RISC-V} core (R2) consumed \SI{0.135}{\watt}. These measurements were obtained under the following environmental and thermal conditions: (i) junction temperature of \SI{26.6}{\degreeCelsius}, (ii) thermal margin of \SI{58.4}{\degreeCelsius} (\SI{4.9}{\watt}), and (iii) ambient temperature of \SI{25}{\degreeCelsius}.

The post-synthesis timing analysis revealed that the core R1 exhibited a longest path delay of \SI{13.142}{\nano\second}, whereas the modified core (R2) showed a reduced path delay of \SI{11.831}{\nano\second}, resulting in faster operation. Consequently, clock periods of \SI{14}{\nano\second} and \SI{12}{\nano\second} were used for simulations of R1 and R2, respectively, to ensure a positive Worst Negative Slack (WNS) in both cases.
Execution time was measured for all inference scenarios as shown in Fig.~\ref{fig:execution}, where Time (ms) is plotted on a logarithmic scale.
While both cores required the same number of clock cycles, their differing clock periods resulted in different execution times, highlighting the performance gain from the reduction.
We evaluated inference strategies T1 and T2 on six datasets (Table~\ref{tab:dataset_summary}) using both the cores R1 and R2. For each dataset, we analyzed the accuracy of the TM model (Table~\ref{tab:accuracy}) across varying numbers of clauses and compared it with the accuracy of a BNN model. 
The resource utilization of the unmodified core (R1) and the reduced core (R2) was analyzed post-synthesis (Table~\ref{tab:resource}).
The number of LUTs was reduced by 28.4\%, and the number of FFs by 12.6\%. The DSP usage saw a 100\% reduction due to the removal of the multiplier and divider units in R2.
TM inference strategies, T1 (Vanilla TM) and T2 (Modified TM), were simulated on both cores R1 and R2 across all clause configurations. Additionally, BNN inference was executed for each dataset on both cores. Using a hardware-software co-design approach, we modified the TM inference to reduce memory usage and improve execution speed. Subsequently, the \mbox{RISC-V} core was streamlined to support a limited set of instructions tailored for TM inference. This co-optimization enabled a maximum reduction of 99.28\% in execution time and an average reduction of 96.55\% across all inference scenarios.
\begin{table}[!t]
\centering
\scriptsize
\caption{Summary of the datasets used in experiments.}

\setlength{\tabcolsep}{2pt}
\renewcommand{\arraystretch}{1.0}

\begin{tabular}{|>{\centering\arraybackslash}p{1.4cm}|
                >{\centering\arraybackslash}p{3.9cm}|
                >{\centering\arraybackslash}p{0.65cm}|
                >{\centering\arraybackslash}p{1.0cm}|
                >{\centering\arraybackslash}p{0.9cm}|}
\hline
\textbf{Dataset} & \textbf{Description} & \textbf{\#Class} & \textbf{\#Features} & \textbf{\#Samples} \\
\hline
CIFAR-2 \cite{cifar2} & 2-class CIFAR animal vs. non-animal & 2 & 324 & 60000 \\
Statlog \cite{statlog} & 3-D and 2-D vehicle images & 4 & 360 & 846 \\
Gesture \cite{gesture} & Hand, wrist, head, and spine positions & 5 & 180 & 9901 \\
Gas \cite{gas_sensor} & Chemical sensor data for 6 gases & 6 & 128 & 13900 \\
EMG \cite{emg_dataset} & EMG data for static hand gestures & 8 & 160 & 14232 \\
FMNIST \cite{FMNIST} & 28×28 grayscale fashion images & 10 & 784 & 70000 \\
\hline
\end{tabular}
\label{tab:dataset_summary}
\end{table}

\begin{table}[!t]
\centering
\scriptsize
\caption{Inference accuracy of TM and BNNs (in \%).}

\renewcommand{\arraystretch}{1}
\setlength{\tabcolsep}{2pt}

\begin{tabular}{|l|c|c|c|c|c|c|c|}
\hline
\multicolumn{8}{|c|}{\textbf{Clauses per Class}} \\
\hline
\textbf{Dataset} & 
\textbf{50} & 
\textbf{100} & 
\textbf{150} & 
\textbf{200} & 
\textbf{250} &
\textbf{300} &
\textbf{BNN} \\
\hline
CIFAR-2 \cite{cifar2} & 78.78 & 83.5 & 85.9 & 86.95 & 87.75 & \cellcolor{lightgreen}88.18 & 60.0 \\
Statlog \cite{statlog} & 76.5 & 78.8 & 78.2 & 80.0 & \cellcolor{lightgreen}80.6 & 80.0 & 72.4 \\
Gesture \cite{gesture} & 64.2 & 68.0 & 71.0 & 73.6 & 75.0 & 75.9 & \cellcolor{lightgreen}76.5 \\
Gas \cite{gas_sensor} & 72.9 & 85.0 & 85.5 & 86.3 & \cellcolor{lightgreen}86.9 & \cellcolor{lightgreen}86.9 & 85.8 \\
EMG \cite{emg_dataset} & 80.3 & 84.6 & 85.4 & 86.0 & \cellcolor{lightgreen}86.1 & 85.8 & 83.9 \\
FMNIST \cite{FMNIST} & 82.72 & 84.31 & 84.72 & 84.77 & 84.2 & \cellcolor{lightgreen}84.81 & 80.04 \\
\hline
\end{tabular}
\label{tab:accuracy}
\end{table}

\begin{table}[!t]
\centering
\scriptsize

\caption{Resource utilization comparison.}

\renewcommand{\arraystretch}{1.0}
\setlength{\tabcolsep}{3pt}

\begin{tabular}{|l|c|c|c|c|c|}
\hline
\textbf{Resources} & 
\textbf{Availability} &
\makecell{\textbf{Usage}  \textbf{(R1)}} &
\makecell{\textbf{R1\%} } & 
\makecell{\textbf{Usage}  \textbf{(R2)}} & 
\makecell{\textbf{R2\%} } \\
\hline
LUT & 53200  & 3493 & 6.57\% & 2501 & 4.70\% \\
FF  & 106400 & 2393 & 4.70\% & 2092 & 1.97\% \\
DSP & 200    & 4    & 1.82\% & 0    & 0\%    \\
\hline
\end{tabular}
\label{tab:resource}
\end{table}
\begin{figure*}[htbp]
\centering
\includegraphics[width=0.76\textwidth]{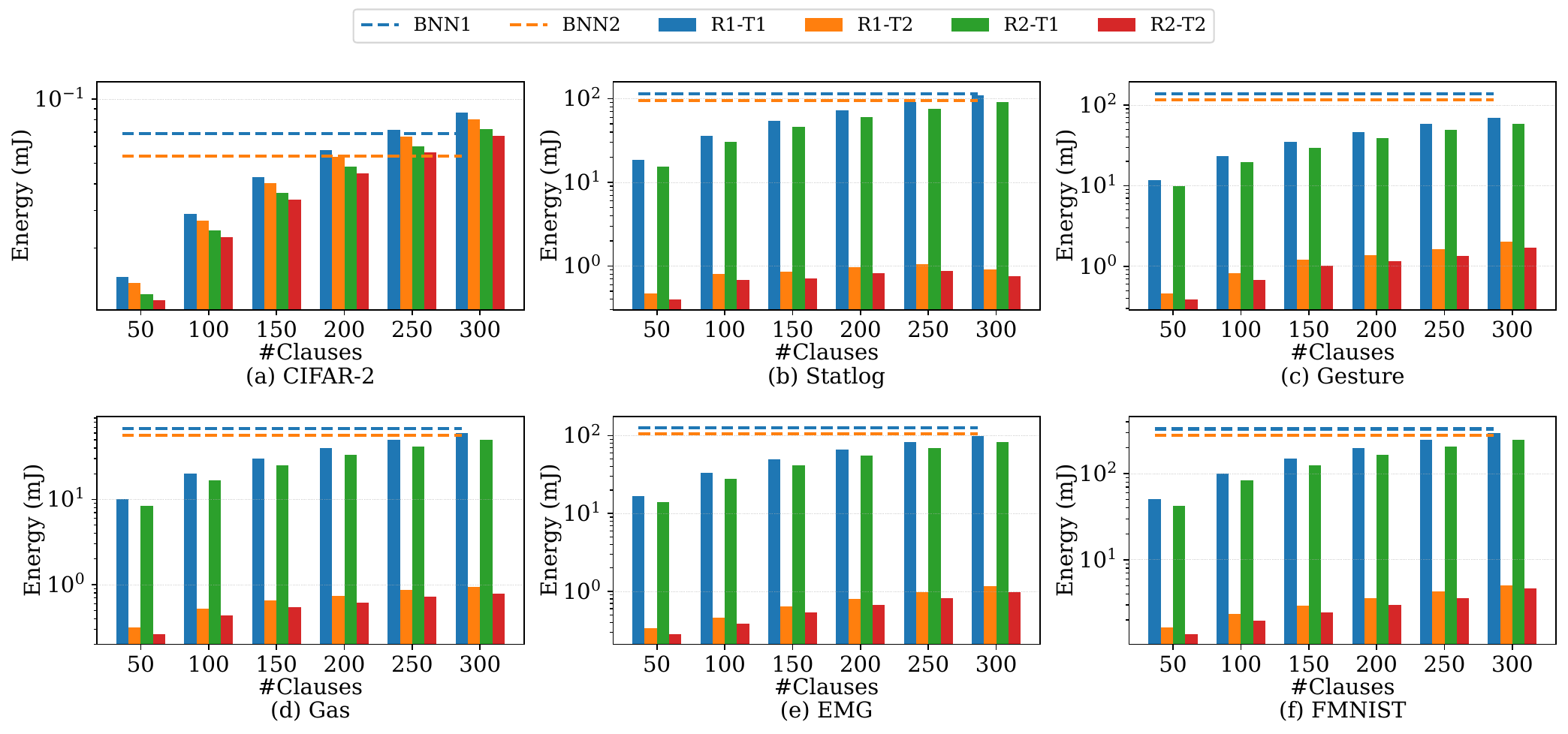}
   \caption{Energy Consumption (in mJ) vs.\ number of clauses per class across datasets.}
\label{fig:energy}
\end{figure*}

\vspace{-0.3cm}
We also observed that for similar accuracy, TM consistently outperforms BNNs in terms of execution time, with the modified TM inference showing a significant performance advantage over BNN.
To estimate the energy consumed during inference, we use the relation $E = P \times T$, where $E$ denotes energy (mJ), $P$ represents total on-chip power (W), and $T$ is the execution time (ms).
The plots in Fig.~\ref{fig:energy} illustrate the variation in energy consumption of TM inference with respect to the number of clauses, with Energy (mJ) plotted on a logarithmic scale.
The dotted horizontal lines indicate the corresponding energy consumption of BNN inference for comparison.
Reducing the instruction set in the core resulted in a power reduction of \SI{3}{\milli\watt} (from \SI{0.138}{\watt} to \SI{0.135}{\watt}), amounting to a 2.18\% decrease in hardware power consumption. When combined with the inference algorithm optimization, we observed an average energy reduction of 96.63\%, with a maximum reduction of 99.3\% across all datasets and clause configurations.
\section{Conclusions and Future Work}
\label{sec:CONCLUSION}

This work addresses the challenge of designing a programmable processor with low energy, fast operation, and reduced overhead for edge AI inference. We presented a domain-specific \mbox{RISC-V} design flow for Tsetlin Machine (TM) inference, guided by instruction profiling and ISA reduction. Results show that TM achieves comparable or higher accuracy than BNNs, with up to 98\% lower execution time and 29.7$\times$ energy savings. This demonstrates the effectiveness of the reduced programmable \mbox{RISC-V} architecture for edge AI.
Future work will explore custom instructions, architectural simplifications, and a design automation flow for RISC-V-based domain-specific processors targeting TM workloads.

\bibliographystyle{IEEEtran}
\bibliography{references_2}

\end{document}